\documentclass{article}

 \usepackage[dblblindworkshop, final]{neurips_2025}

\usepackage[utf8]{inputenc} 
\usepackage[T1]{fontenc}    
\usepackage{hyperref}       
\usepackage{url}            
\usepackage{booktabs}       
\usepackage{amsfonts}       
\usepackage{nicefrac}       
\usepackage{microtype}      
\usepackage{xcolor}         

\usepackage{amsmath,amsfonts,bm}
\usepackage{amsthm} 
\usepackage{mathtools}
\usepackage{graphicx} 
\usepackage{wrapfig} 
\usepackage{subfigure} 
\usepackage{caption}
\usepackage{booktabs}       %

\workshoptitle{Efficient Reasoning}

\title{Optimizing Reasoning Efficiency through Prompt Difficulty Prediction}

\author{%
  Bo Zhao\thanks{Work done during an internship at Capital One. Contact: \texttt{bozhao@ucsd.edu}.} \\
  UC San Diego\\
  \And
  Berkcan Kapusuzoglu \\
  Capital One\\
  \And
  Kartik Balasubramaniam \\
  Capital One\\
  \AND
  Sambit Sahu \\
  Capital One\\
  \And
  Supriyo Chakraborty \\
  Capital One\\
  \And
  Genta Indra Winata \\
  Capital One\\
}

\newcommand{\BK}[1]{{\textcolor{cyan}{\footnotesize\sf[BK: #1]}}}

\renewcommand{\BK}[1]{}

\begin{document}

\maketitle

\begin{abstract}
Reasoning language models perform well on complex tasks but are costly to deploy due to their size and long reasoning traces.
We propose a routing approach that assigns each problem to the smallest model likely to solve it, reducing compute without sacrificing accuracy.
Using intermediate representations from s1.1-32B, we train lightweight predictors of problem difficulty or model correctness to guide routing across a pool of reasoning models.
On diverse math benchmarks, routing improves efficiency over random assignment and matches s1.1-32B’s performance while using significantly less compute.
Our results demonstrate that difficulty-aware routing is effective for cost-efficient deployment of reasoning models.
\BK{Is the improvement statistically significant?, How does this compare to confidence-based routing or other baselines?}
\end{abstract}

\section{Introduction}

Recent advances in large language models (LLMs) have significantly improved reasoning across math, science, and general problem-solving tasks \citep{li2025system}. However, these gains come with high computational costs, especially when large models are used uniformly across tasks of varying difficulty. Many problems can be solved by smaller models at a fraction of the cost, suggesting that adaptive model selection could greatly improve efficiency \citep{openai2025introducinggpt5}.

This paper investigates whether intermediate representations from LLMs can be used to predict problem difficulty and guide model selection. We train lightweight classifiers to predict either the difficulty of a problem or the likelihood that a model will answer it correctly. These predictors, trained on outputs from the s1.1-32B model, enable routing strategies that assign each problem to the smallest model expected to succeed.
Prior work relied on problem-specific, hand-designed metrics (e.g., number of obstacles in a maze) to assess difficulty \citep{saha2025system}. In contrast, we propose a learned, general-purpose classifier for labeling problem difficulty.

We evaluate this approach across multiple reasoning benchmarks, comparing against baselines that use a single model or assign models at random. Our results show that mid-layer LLM representations are highly informative for difficulty and correctness prediction, and that routing based on these signals can dramatically reduce inference cost while maintaining high accuracy of the large models.


\section{Related Work}




\paragraph{Improving efficiency using prompt attributes}
Prediction of expected performance \citep{shnitzer2023large, lu2023routing} or uncertainty level \citep{chuang2025learning, chuang2025confident}, often combined with the goal to minimize cost \citep{mohammadshahi2024routoo, hu2024routerbench, ding2024hybrid, vsakota2024fly}, has been used to route query to LLMs with different expertise or size.
As reasoning models emerge recently, people use similar query attributes to route to models with different reasoning capabilities to mitigate overthinking \citep{pu2025thoughtterminator, saha2025system}. 
We contribute to this line of work through developing a prediction model for fine-grained analysis of query attributes and automated routing. 

\paragraph{Controllable fast and slow thinking}
Balancing between fast and slow thinking can be achieved by combining models with different capabilities or exploring shortcuts in the reasoning trace \citep{sui2025stop}.
System-1.x Planner \citep{saha2025system} decomposes problems and routes sub-tasks to fast or slow solvers based on a task-based hardness function. 
The Fast-Slow-Thinking framework \citep{sun2025fast} first simplifies a task by removing constraints for a quick solution, then refines it by reintroducing them. 
Dualformer \citep{su2024dualformer} trains a single model on randomized reasoning traces with different parts dropped, enabling it to learn both full reasoning and cognitive shortcuts. 
System-1.5 Reasoning \citep{wang2025system} allocates computational effort by learning shortcuts along model depth in latent-space reasoning.




\section{Methods and Experiment Settings}

\begin{figure}[t]
\begin{center}
\includegraphics[width=0.9\columnwidth]{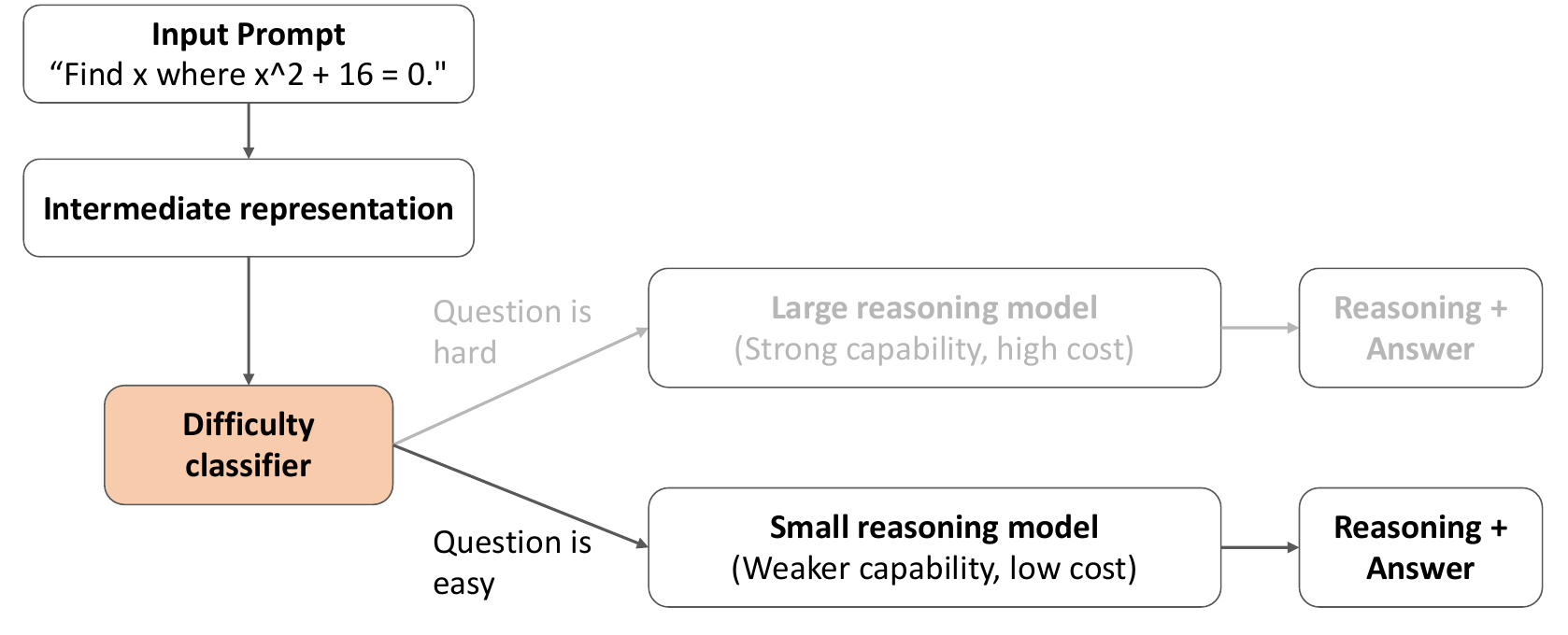}
\caption{A classifier predicts problem difficulty from intermediate representations, and route each problem to the smallest reasoning model likely to solve it. This reduces inference cost while maintaining accuracy.}
\label{fig:overview}
\vspace{-5pt}
\end{center}
\end{figure}

We train a model to predict properties of reasoning problems and route each to an appropriately sized reasoning model (Figure \ref{fig:overview}).
We assume access to a pool of reasoning models with varying capacity and inference cost.
We first use intermediate outputs from s1.1-32B to train a predictor of either problem difficulty or model correctness.
This predictor then guides routing, assigning each question to the smallest model likely to solve it.
We evaluate the router by measuring accuracy and inference time against a baseline of random assignment.

\subsection{Difficulty Level Prediction}
\label{sec:difficulty-level-prediction}
We train a model to predict question difficulty using the MATH dataset \citep{hendrycks2021measuring}, which includes 7500 competition problems labeled from 1 (easiest) to 5 (hardest).
We split the dataset into 6000 for training and 1500 for validation. 
For the difficulty level predictor, we train a 3-layer MLP with s1.1-32B's layer outputs as input (dimension 5120), hidden dimensions 256 and 64, and cross entropy loss.
We train 20 epochs using batch size 32 and learning rate $10^{-5}$.

\subsection{Model Accuracy Prediction}
\label{sec:model-accuracy-prediction}
To predict whether a given LLM can solve a problem, we construct the MathCombined dataset, which consists of 3136 reasoning tasks with ground truth solution but without difficulty levels. 
This dataset includes problems from AIME24 \citep{maa2024aime}, AMC23 \citep{maa2023amc}, GSM8k \citep{cobbe2021gsm8k}, Minerva \citep{lewkowycz2022solving}, OlympiadBench \citep{he2024olympiadbench}, and TheoremQA \citep{chen2023theoremqa}.
We split the data into 1882 for training, 626 for validation, and 628 for router evaluation (Section \ref{sec:router-accuracy}). 
For the model accuracy predictor, we train a 4-layer MLP with s1.1-32B's layer outputs as inputs (dimension 5120), hidden dimensions 8192, 2048, and 128, and binary cross entropy loss.
We train 20 epochs using batch size 32 and learning rate $5 \times 10^{-6}$.

We aim to predict correctedness for the following models: Mixtral-8x7B-instruct \citep{jiang2024mixtral}, OLMo-2-1124-7B-Instruct \citep{olmo20242}, Llama-3.1-8B-Instruct \citep{grattafiori2024llama}, phi-4 \citep{abdin2024phi4}, Llama-3.1-Nemotron-Nano-8B \citep{bercovich2025llama}, Llama-3.3-70B-Instruct \citep{grattafiori2024llama}, Llama-3.3-Nemotron-Super-49B \citep{bercovich2025llama}, and s1.1-32B \citep{muennighoff2025s1}.
Additional dataset and model details can be found in Appendix \ref{appendix:dataset} and \ref{appendix:models}.

\subsection{Router Based on Difficulty Prediction} 
We design a router using the difficulty level predictor from Section \ref{sec:difficulty-level-prediction}. 
A problem is assigned to a larger model if the predicted difficulty exceeds a threshold, and to a smaller model otherwise.
The router is evaluated on the evaluation split of MathCombined.

\begin{figure}[t]
\begin{center}
\ \ \ (a) \hfill (b) \hfill ~ \\
\includegraphics[width=0.38\columnwidth]{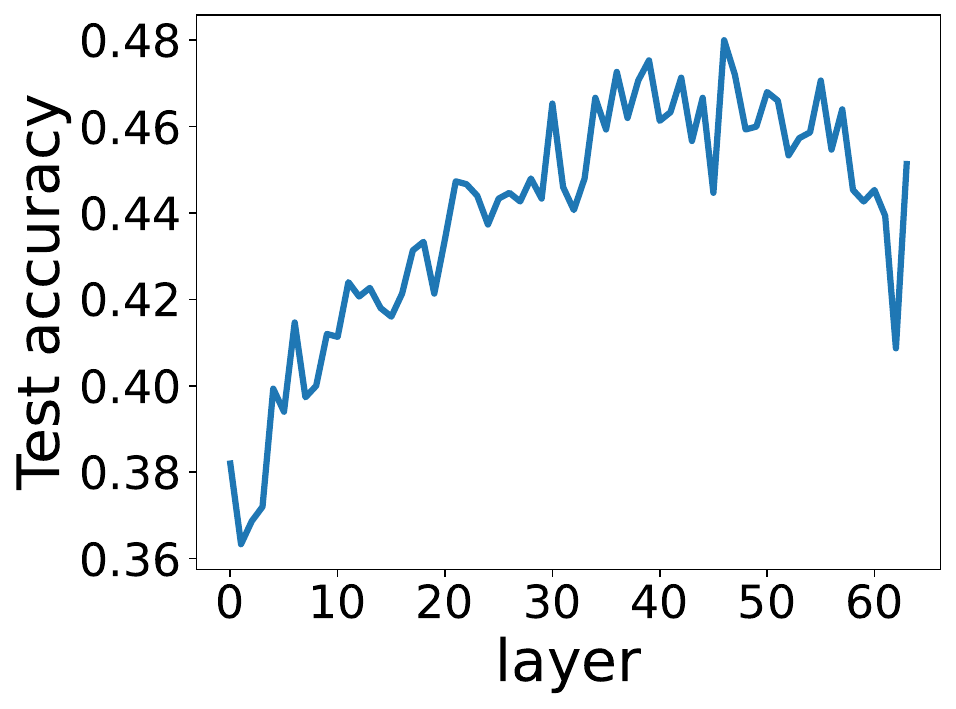}
\hspace{30pt}
\includegraphics[width=0.38\columnwidth]{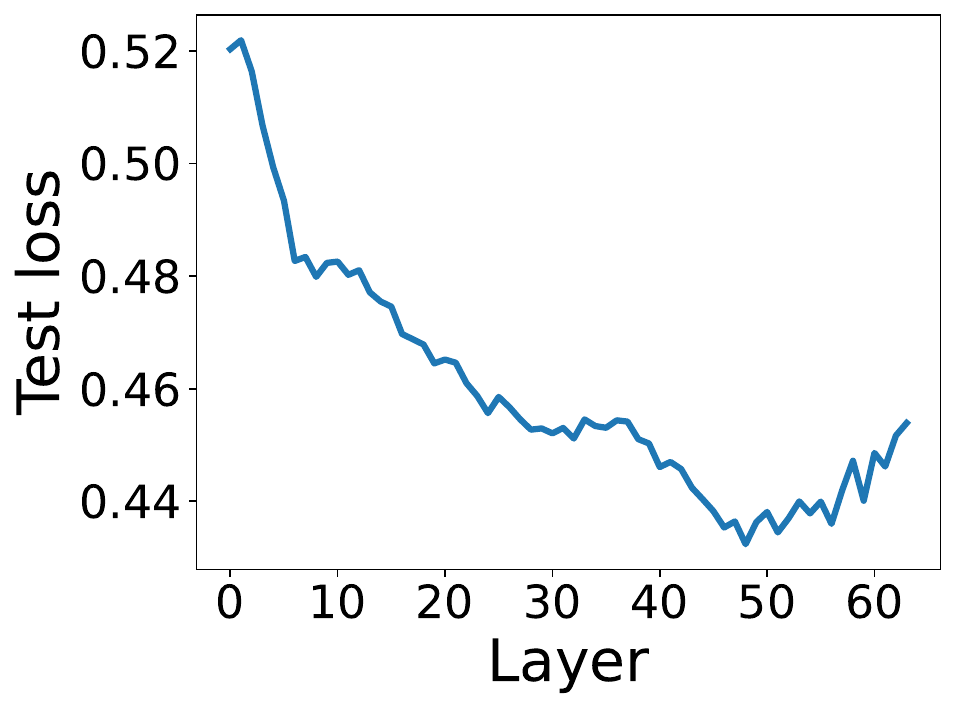}
\caption{Prediction performance using outputs from different layers of s1.1-32B on (a) question difficulty level and (b) whether various language models can answer the given question correctly. Middle layers provide the most informative representations.}
\label{fig:accuracy-vs-layer-depth-s1}
\end{center}
\end{figure}

Using the embeddings from a smaller model (e.g. Llama Nemotron 8B), one can still train a decent difficulty predictor and router. Similar predictor and router performance using Llama Nemotron 8B's layers can be found in Appendix \ref{appendix:llama-nemotron}.

\begin{figure}[t]
\begin{center}
\includegraphics[width=0.325\columnwidth]{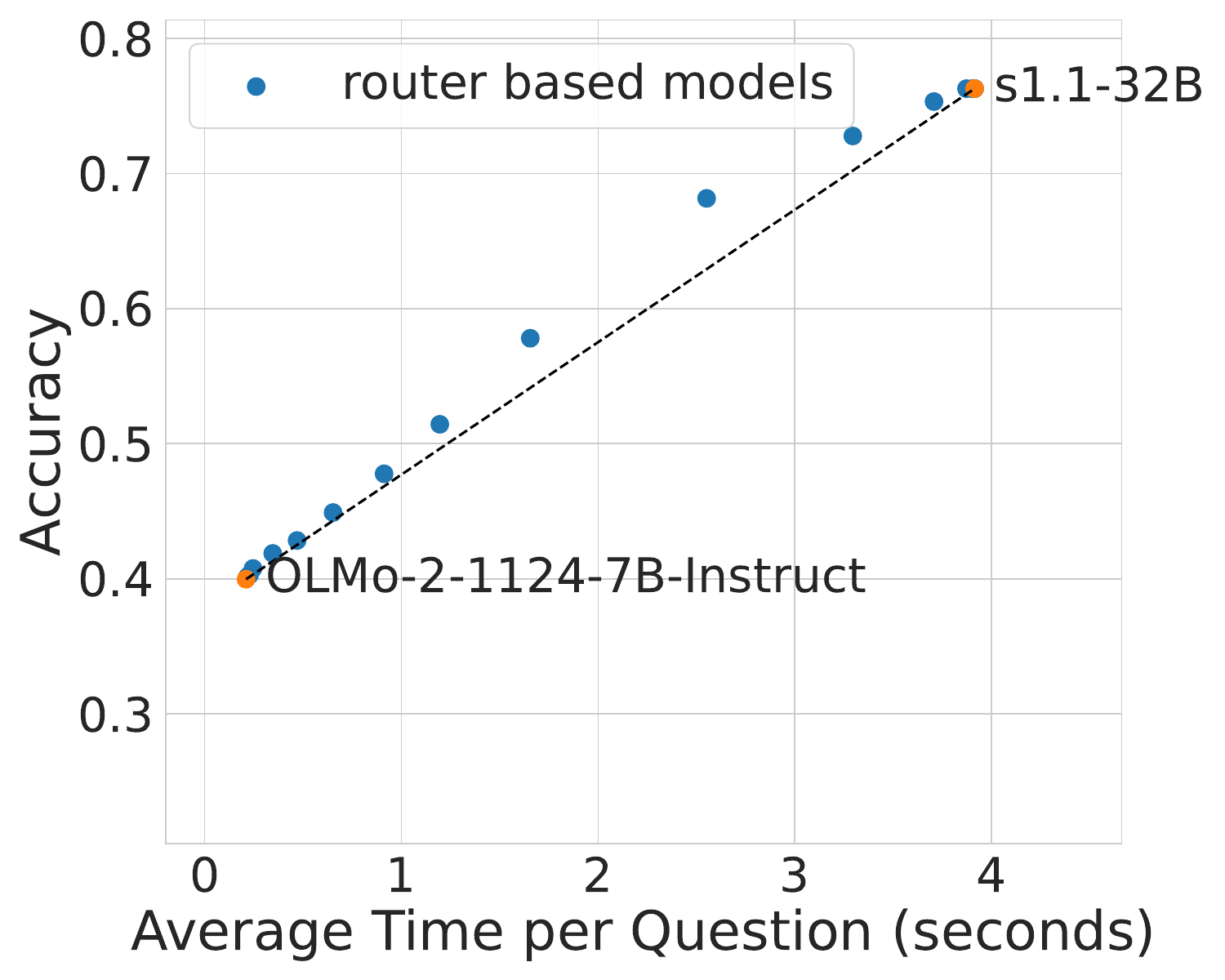}
\includegraphics[width=0.325\columnwidth]{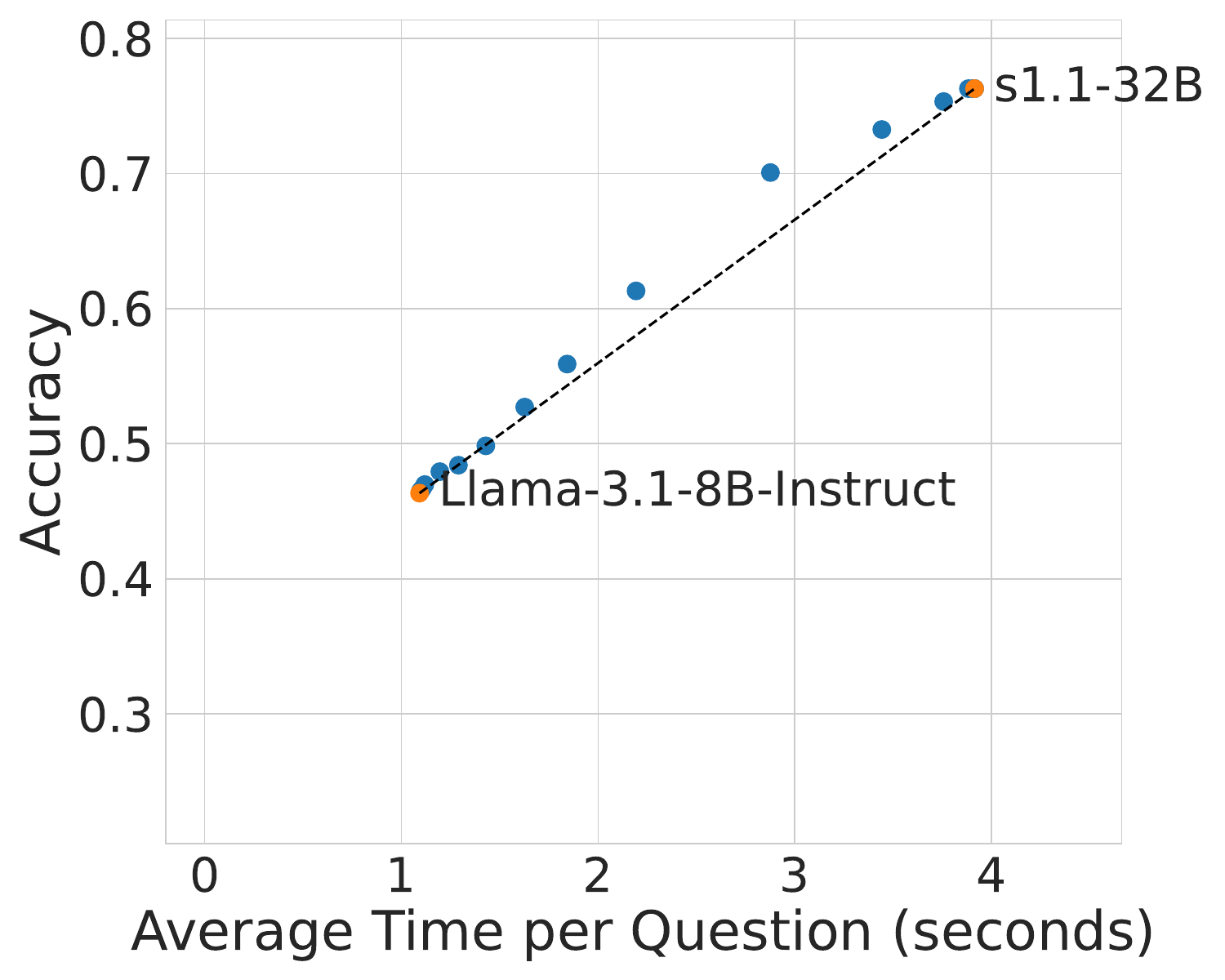}
\includegraphics[width=0.325\columnwidth]{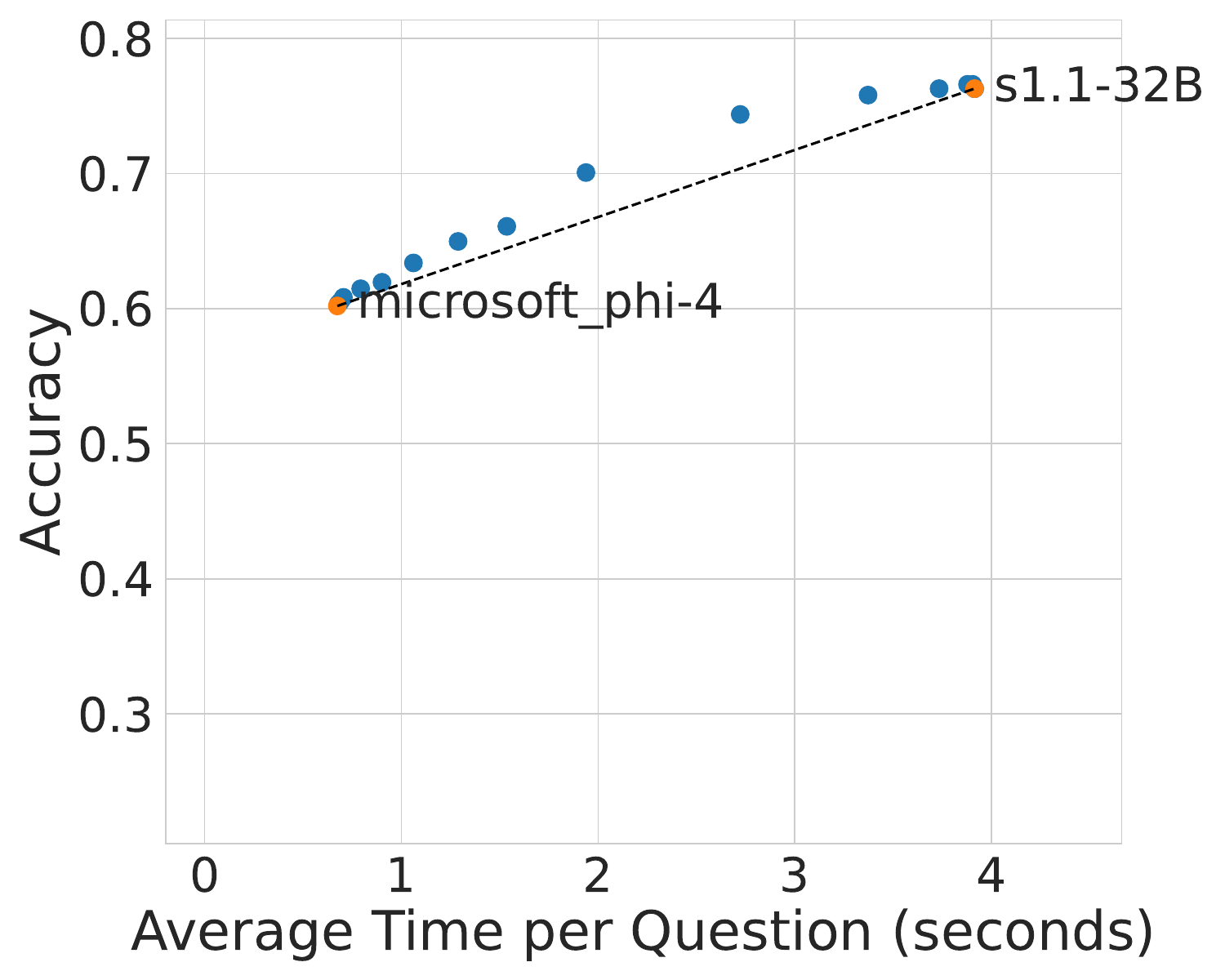}
\caption{Performance of difficulty-based routing using s1.1-32B layer outputs. A problem is routed to a larger model if the predicted difficulty exceeds a threshold, and to a smaller model otherwise. Blue dots indicate router-based systems with thresholds between 2.1 and 2.9; orange dots show baseline models. Routers consistently outperform random assignment.}
\label{fig:route-using-difficulty-s1}
\end{center}
\end{figure}

\begin{figure}[t]
\begin{center}
\includegraphics[width=0.77\columnwidth]{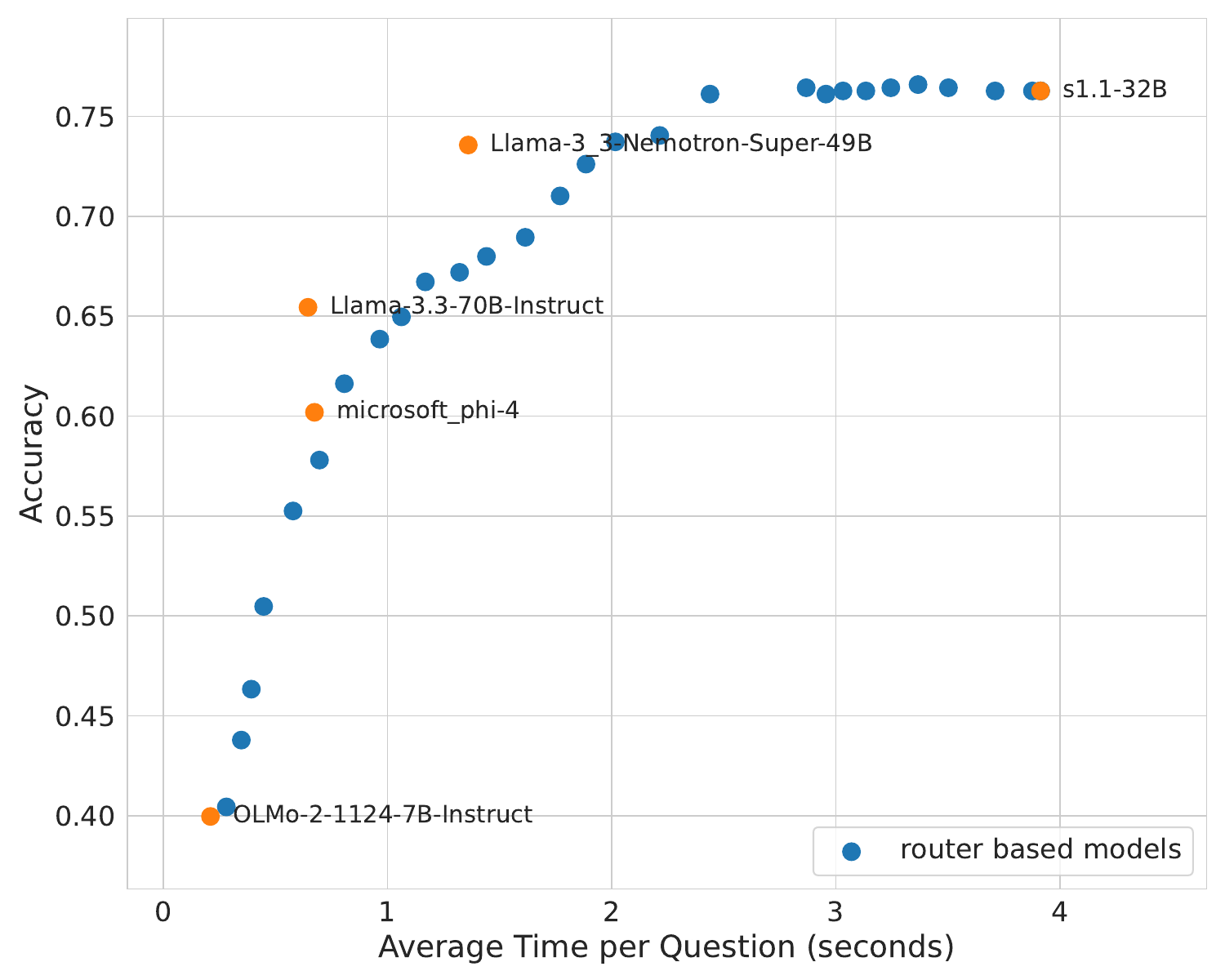}
\caption{Performance of accuracy-based routing using s1.1-32B layer outputs. Each problem is routed to the weakest model with predicted correctness above a threshold. Blue dots correspond to thresholds between 0.05 and 0.9.}
\label{fig:router-using-accuracy-s1}
\vspace{-5pt}
\end{center}
\end{figure}

\subsection{Router Based on Model Accuracy Prediction} 
\label{sec:router-accuracy}
We build another router using the model accuracy predictor from Section \ref{sec:model-accuracy-prediction}. 
Each problem is assigned to the smallest model whose predicted correctness exceeds a threshold. 
The router is evaluated on the evaluation split of MathCombined.

Finally, while ideally we would leverage intermediate representations from all candidate models to capture model-specific interpretations of problem difficulty, we adopt a pragmatic approach using embeddings from a single representative model (S1.1-32B). 
This approach assumes that difficulty patterns captured by one sufficiently capable model can generalize across similar model architectures.

\section{Results}

Figure \ref{fig:accuracy-vs-layer-depth-s1} shows prediction performance when using outputs from different layers of s1.1-32B to estimate either problem difficulty or model correctness.
We find that middle layers yield the best prediction models, which suggests that middle layers contain more information relevant to problem difficulty.
This aligns with previous findings that intermediate layers often have better performance in downstream tasks than the final layer \citep{skean2025layer}.
\BK{Why middle layers specifically? (for later, it would be great if we could provide theoretical justification, e.g., middle layers balance low-level features with high-level reasoning etc.) Test across multiple model architectures}
Accordingly, we use layer 45 and the prediction models trained on it for the routers.

\BK{No error bars, or confidence intervals. Maybe we can provide more statistical analysis on the results}

The router based on these prediction models yields reasoning models with higher overall efficiency.
Figure \ref{fig:route-using-difficulty-s1} shows performance of router based on difficulty prediction. With a router, we are able to obtain models that outperforms randomly assigning between the two given models.
Figure \ref{fig:router-using-accuracy-s1} shows performance of router based on model accuracy prediction. 
With the appropriate thresholds, our proposed model achieves comparable and even slightly better performance as s1.1-32B, while requiring only about two-thirds of the inference compute.
\BK{this may need some rigorous analysis. Also another interesting thing is how does efficiency scale with different model size ratios?}

\section{Discussion}
We proposed a difficulty classifier to improve inference efficiency on reasoning tasks by routing problems to the smallest model likely to solve them.
Beyond efficiency, such classifiers have broader utility. They can provide automatic difficulty annotations for datasets, support curriculum learning and evaluation, and enable selective abstention when a problem is predicted to be too hard.
Future work include improving the classifiers by analyzing the embedding space. 
For example, one could explore routing based on similarity in this space, such as assigning a problem to the model that has successfully solved its nearest neighbors. 
More generally, integrating better routing strategies and extending beyond a single representative model may further enhance efficiency and robustness.

\clearpage


\bibliography{neurips_2025}
\bibliographystyle{plainnat}

\clearpage
\appendix

\section{Additional Dataset Details}
\label{appendix:dataset}

The MathCombined dataset, which consists of 3136 reasoning tasks with ground truth solution but without difficulty levels, includes the following data:
\begin{itemize}
    \item AIME24 \citep{maa2024aime} (30 problems from the 2024 AIME I and AIME II tests)
    \item AMC23 \citep{maa2023amc} (40 problems from the 2023 AMC 12)
    \item GSM8k \citep{cobbe2021gsm8k} (1319 grade school math problems)
    \item Minerva \citep{lewkowycz2022solving} (272 STEM problems at the undergraduate level)
    \item OlympiadBench \citep{he2024olympiadbench} (675 challenging math problems)
    \item TheoremQA \citep{chen2023theoremqa} (800 question-answering covering theorems from math, physics, EE\&CS, and finance)
\end{itemize}

\begin{figure}[h!]
\begin{center}
\includegraphics[width=0.65\columnwidth]{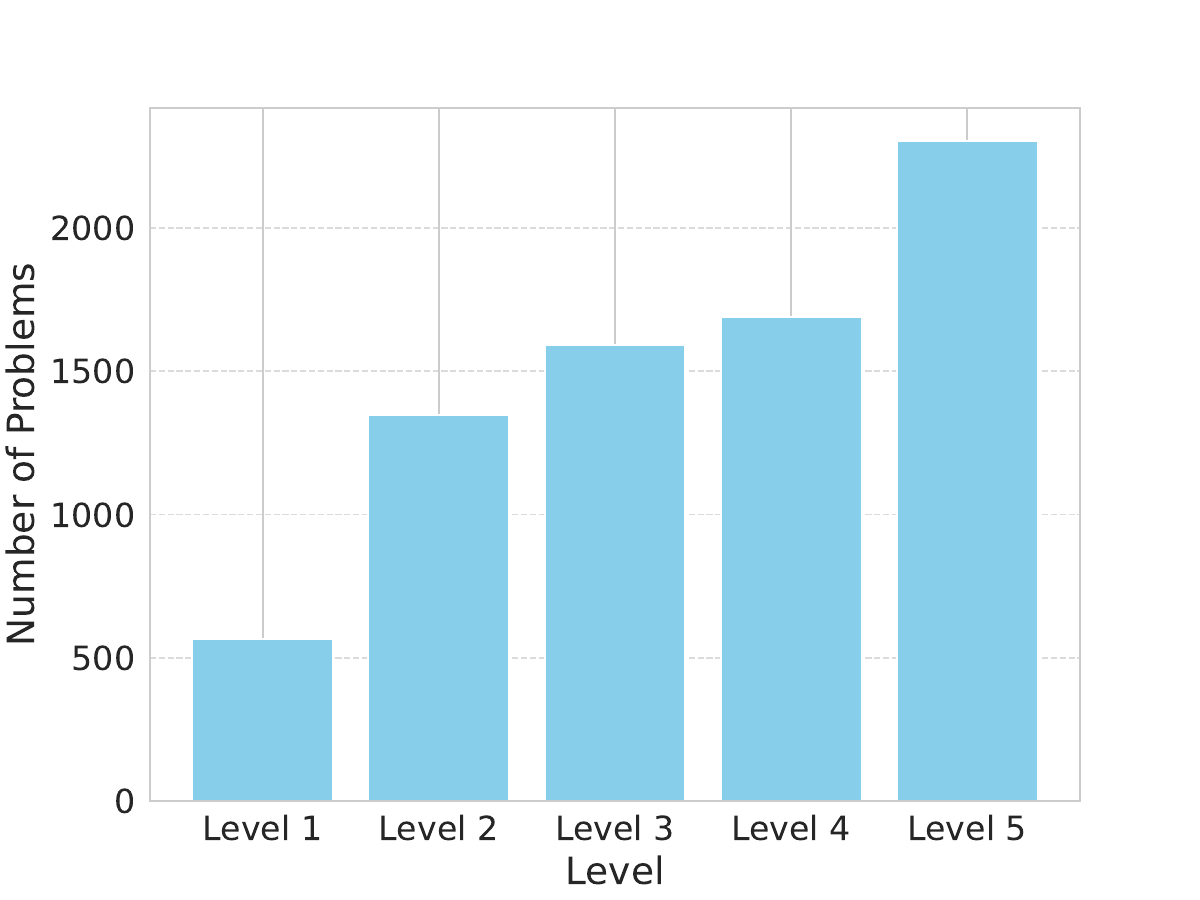}
\caption{Difficulty level distribution of the MATH dataset.}
\label{fig:math-level-distribution}
\end{center}
\end{figure}

The MATH dataset \citep{hendrycks2021measuring} consists of 7500 problems from various mathematics competitions, together with the difficulty level of each problem. 
Figure \ref{fig:math-level-distribution} shows the distribution of difficulty levels in the MATH dataset.

\section{Additional Model Details}
\label{appendix:models}
\begin{figure}[h]
\begin{center}
\includegraphics[width=0.80\columnwidth]{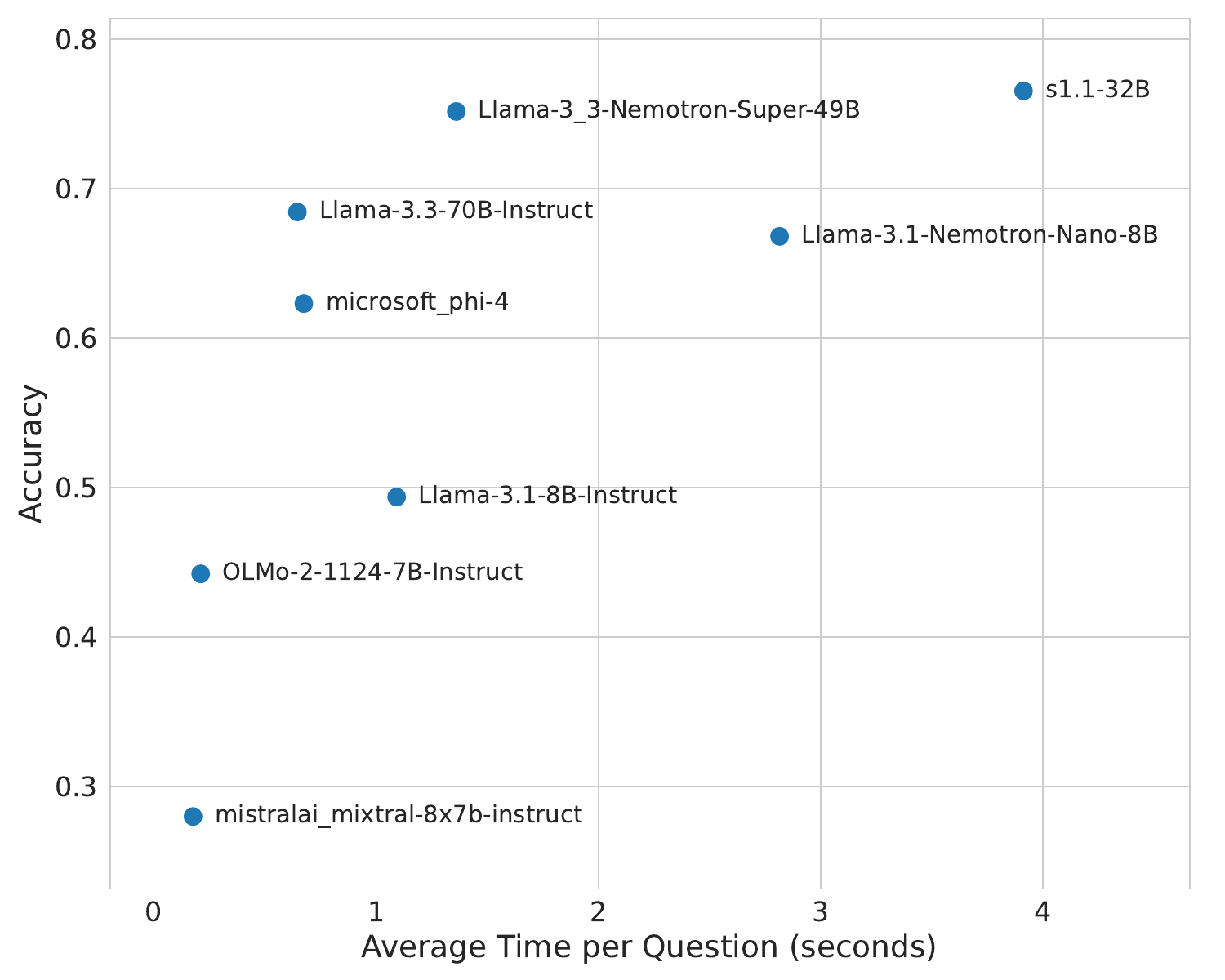}
\caption{Comparison of models' performance on MathCombined. }
\label{fig:model_performance_scatter_plot_all}
\end{center}
\end{figure}

Figure \ref{fig:model_performance_scatter_plot_all} shows the percentage of questions answered correctly versus the average inference time per problem, for all models on the MathCombined dataset.
As a high level trend, models achieving higher accuracy tend to need longer inference time. 
Notable exceptions are Mixtral-8x7B-instruct and Llama-3.1-8B-Instruct, both of which are not reasoning models, as well as Llama-3.1-Nemotron-Nano-8B.
We therefore remove them from the set of models available to the routers.

\begin{figure}[h]
\begin{center}
\includegraphics[width=1.0\columnwidth]{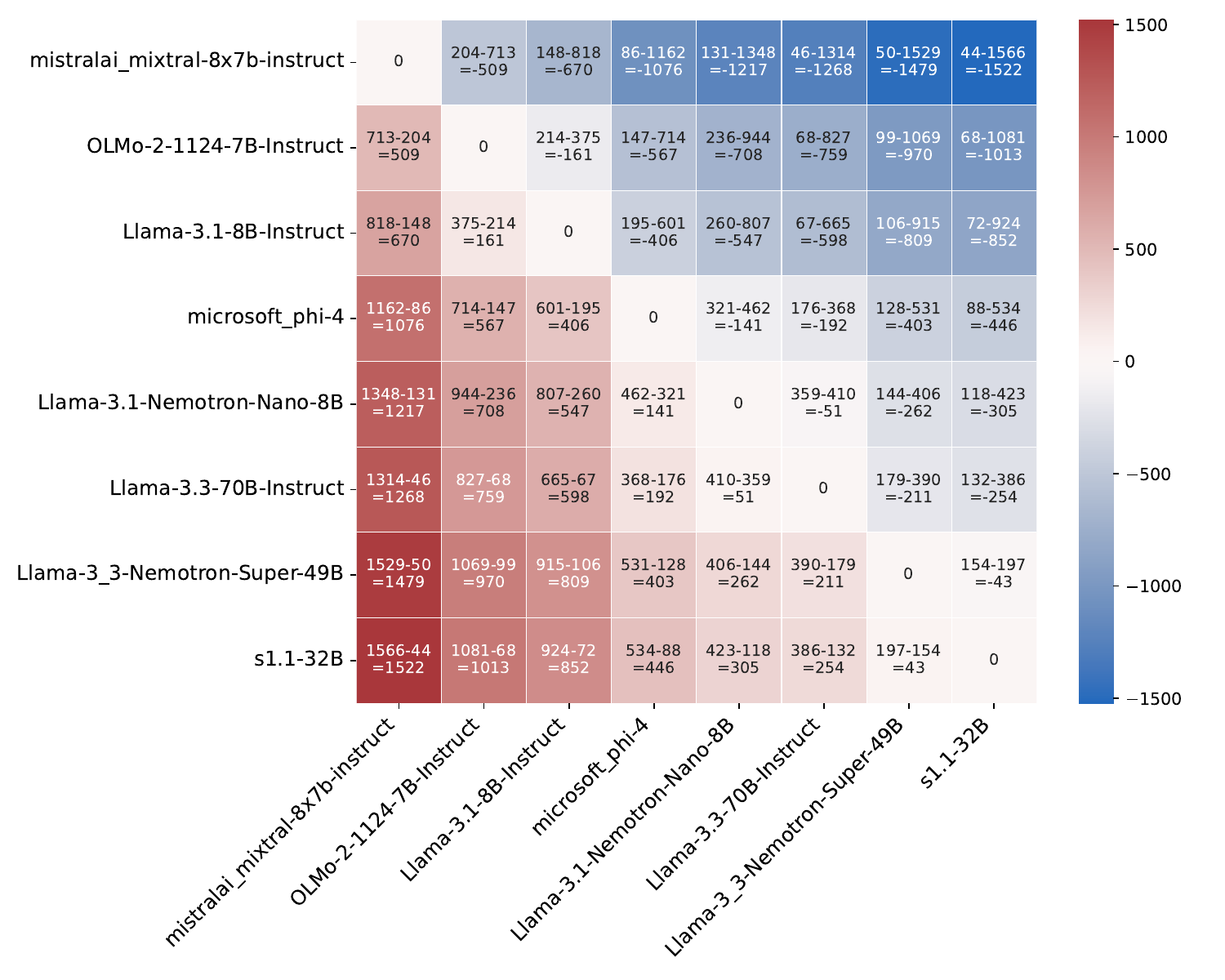}
\caption{Heatmap comparing models' performance on MathCombined. Label[$i$][$j$]: total number of questions model $i$ answers correctly but model $j$ does not $-$ total number of questions model $j$ answers correctly but but model $i$ does not = difference.}
\label{fig:model-correctness-heatmap}
\end{center}
\end{figure}

Figure \ref{fig:model-correctness-heatmap} provides a more fine grained comparison between all models. 
For each pair of models $i, j$, we examine the number of questions model $i$ answers correctly but model $j$ does not. 
Interestingly, for all pairs of models, there are a nonzero number of questions that the first model can answer but the second cannot. 
Even the strongest model (s1.1-32B) fails at a few problems that the weakest model (Mixtral-8x7B-instruct) answers correctly.
This justifies the possibility that our router based model can achieve higher accuracy than always using the strongest model.

\clearpage
\section{Llama Nemotron 8B Results}
\label{appendix:llama-nemotron}

In the main paper, we used the embeddings from s1.1-32B to train the difficulty predictor. 
We show below that using the embeddings from a smaller model, one can still train a decent difficulty predictor and router. 
Below, we show predictor and router performance using Llama Nemotron 8B. 
The routers use the prediction models trained on embeddings from layer 20, which has one of the strongest performance when used to predict problem difficulty.

\begin{figure}[h]
\begin{center}
\ \ \ (a) \hfill (b) \hfill ~ \\
\includegraphics[width=0.4\columnwidth]{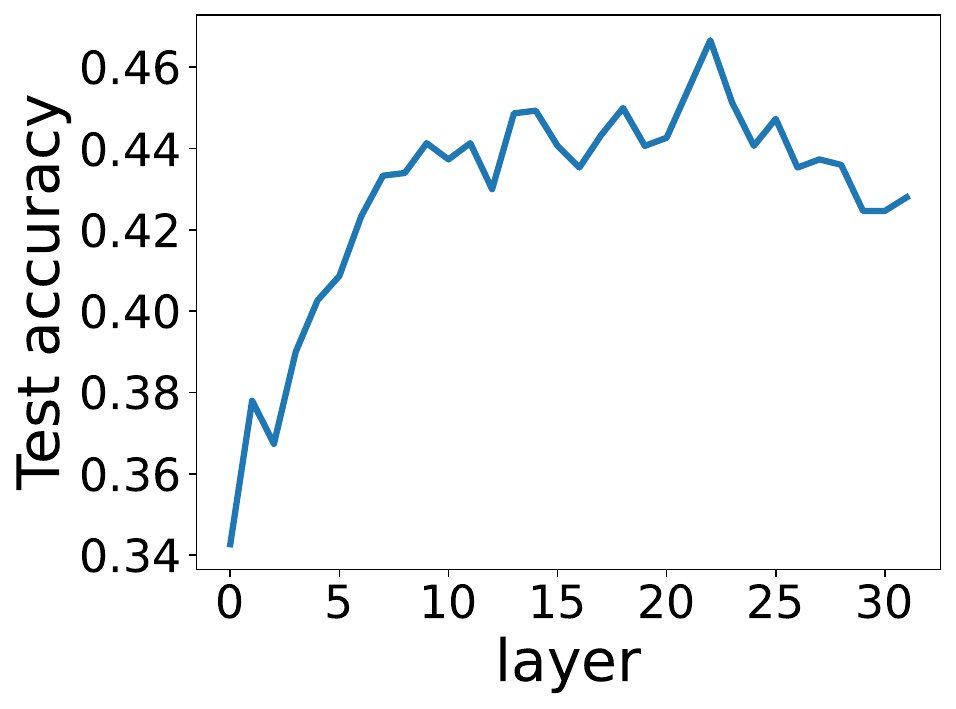}
\hspace{30pt}
\includegraphics[width=0.4\columnwidth]{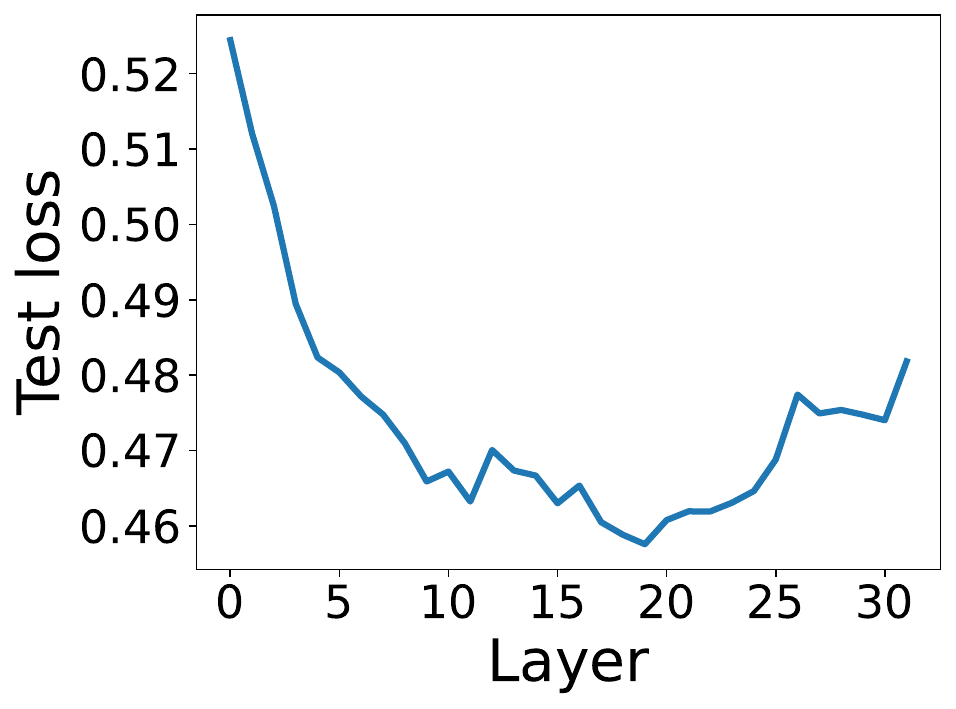}
\caption{Prediction performance using outputs from different layers of Llama-3.1-Nemotron-Nano-8B-v1 on (a) question difficulty level and (b) whether various language models can answer the given question correctly. Middle layers provide the most informative representations.}
\label{fig:accuracy-vs-layer-depth-llama-nemotron}
\end{center}
\end{figure}

\begin{figure}[h]
\begin{center}
\includegraphics[width=0.325\columnwidth]{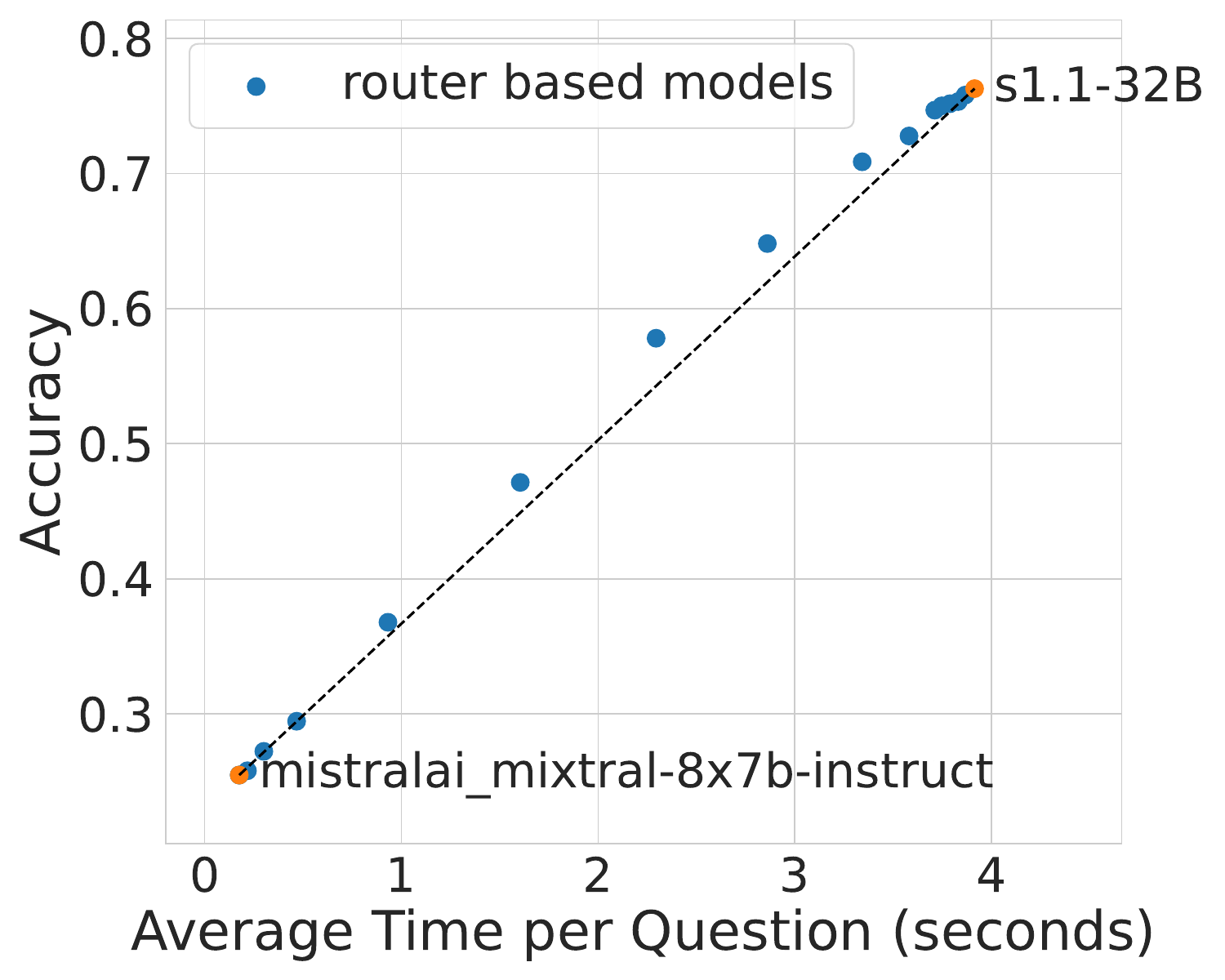}
\includegraphics[width=0.325\columnwidth]{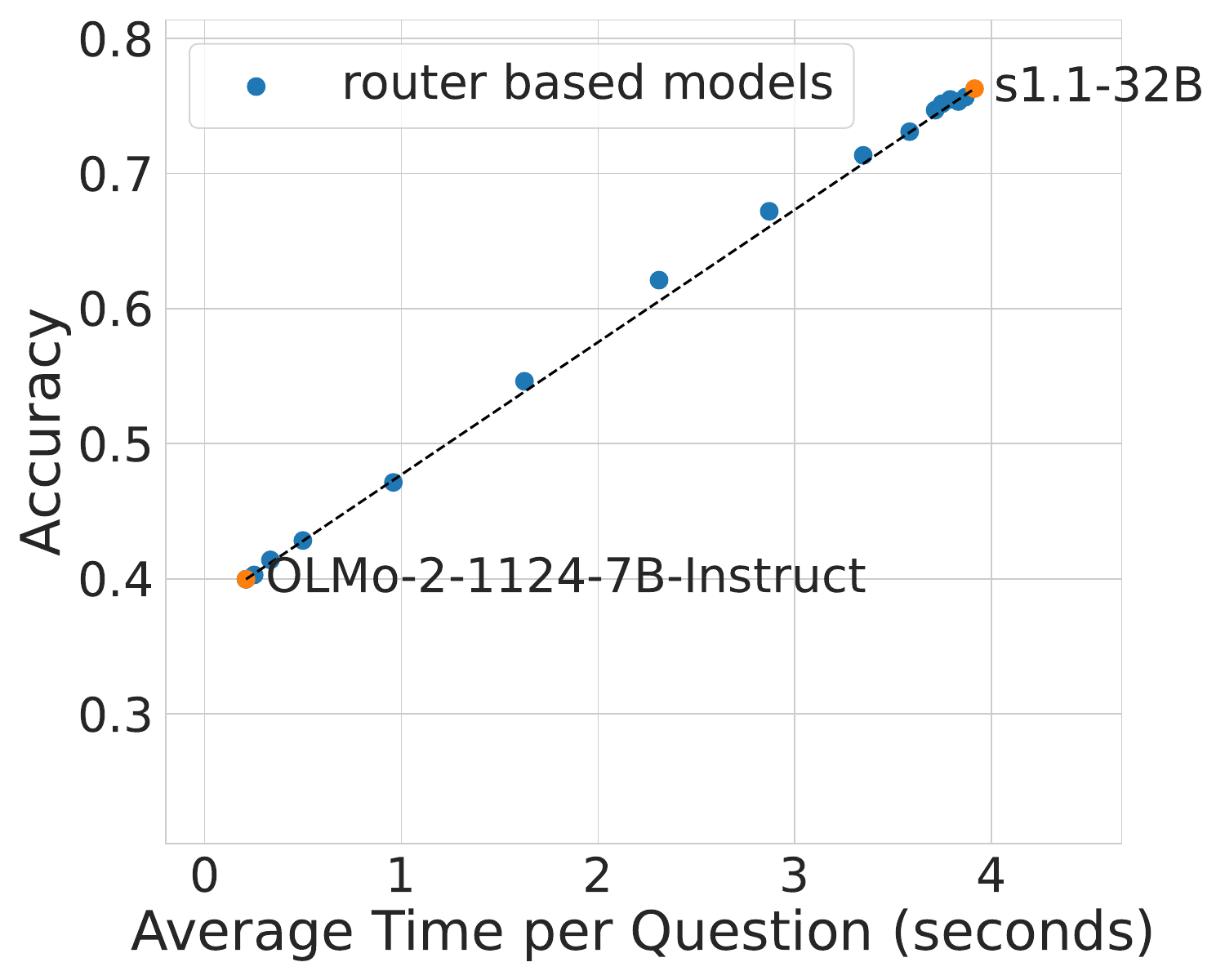}
\includegraphics[width=0.325\columnwidth]{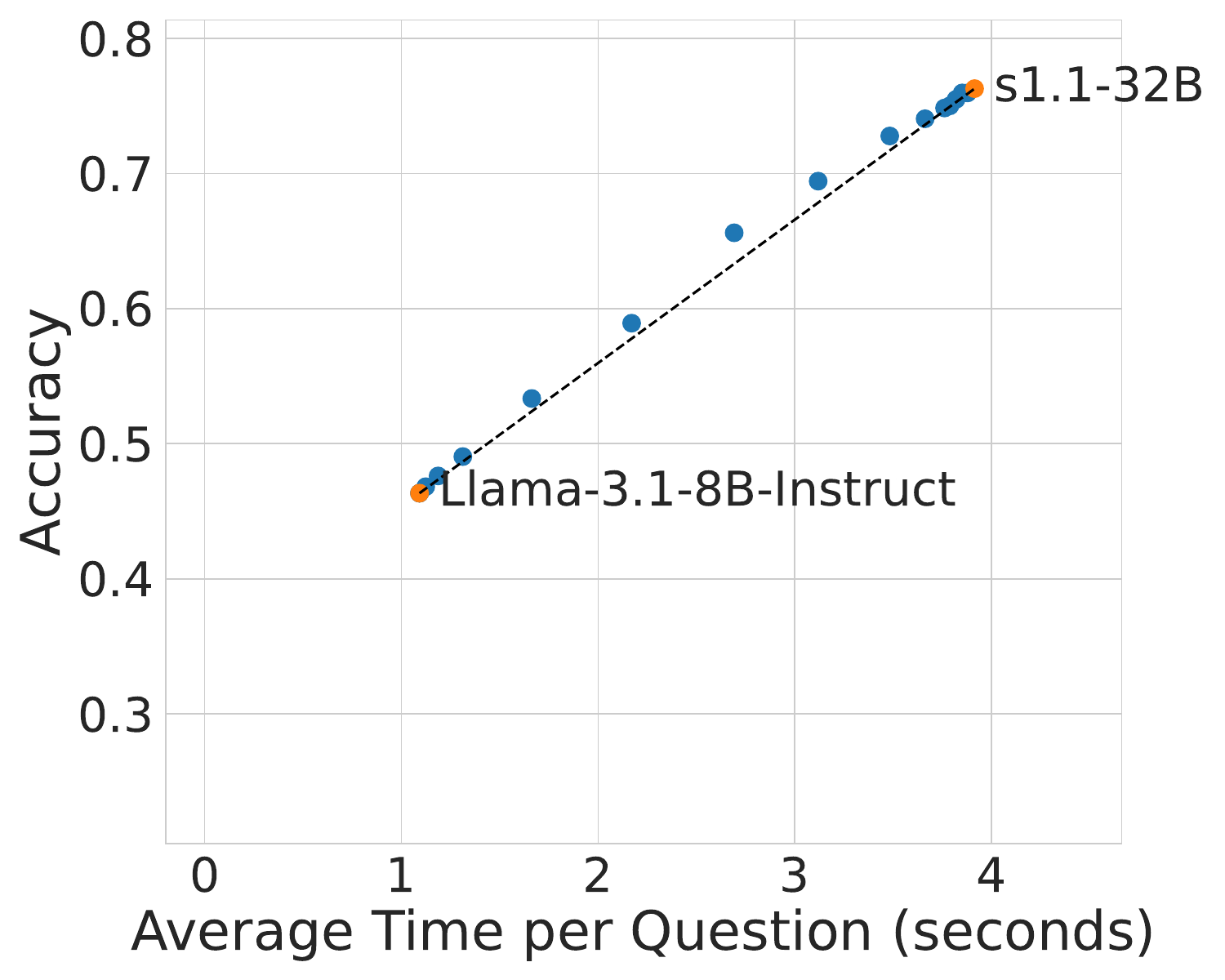}
\caption{Performance of difficulty-based routing using Llama-3.1-Nemotron-Nano-8B-v1 layer outputs. A problem is routed to a larger model if the predicted difficulty exceeds a threshold, and to a smaller model otherwise. Blue dots indicate router-based systems with thresholds between 2.1 and 2.9; orange dots show baseline models. Routers consistently outperform random assignment.}
\label{fig:route-using-difficulty-llama}
\end{center}
\end{figure}

\begin{figure}[h]
\begin{center}
\includegraphics[width=0.80\columnwidth]{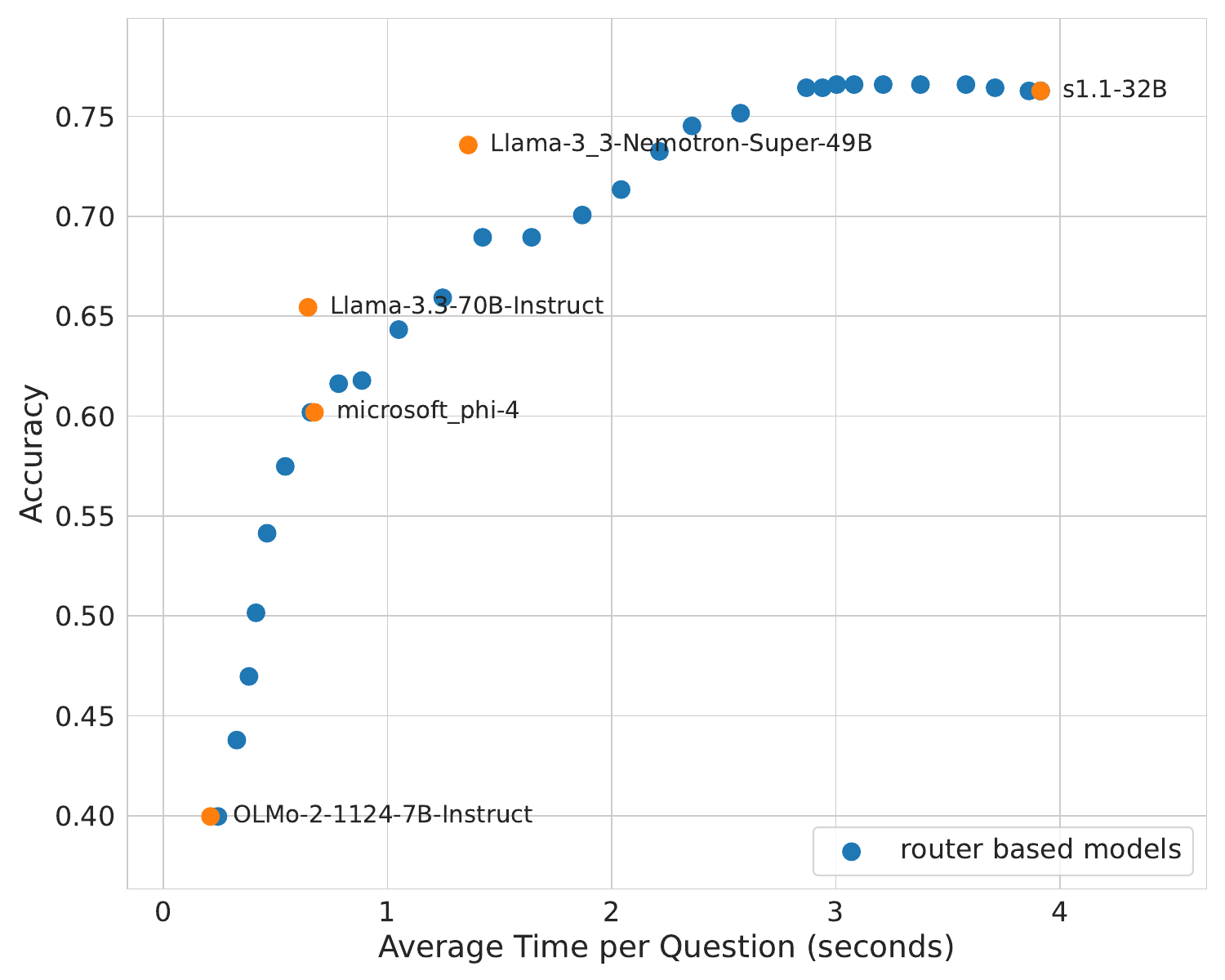}
\caption{Performance of accuracy-based routing using Llama-3.1-Nemotron-Nano-8B-v1 layer outputs. Each problem is routed to the weakest model with predicted correctness above a threshold. Blue dots correspond to thresholds between 0.05 and 0.9.}

\label{fig:router-using-accuracy-llama}
\end{center}
\end{figure}


\end{document}